\documentclass[manuscript,screen]{acmart}

\newcommand{\blue}[1]{{\color[rgb]{0,0.378,0.96} {#1}}}
\AtBeginDocument{%
  }

\begin{document}

\title{Sequential Design of Genetic Circuits Under Uncertainty With Reinforcement
Learning}




© 2026 Michal Kobiela, Diego A. Oyarzún, Michael U.Gutmann. This is the author's version of the
work. It is posted here for personal use. Not for redistribution. This manuscript has been submitted for
publication. \\

\author{Michal Kobiela}
\email{M.Kobiela@sms.ed.ac.uk}
\orcid{0009-0004-9308-1993}
\affiliation{%
  \institution{School of Informatics, The University of Edinburgh}
  \city{Edinburgh}
  \country{UK}
}

\author{Diego A. Oyarzún}
\orcid{0000-0002-0381-5278}
\email{d.oyarzun@ed.ac.uk}
\affiliation{%
  \institution{School of Informatics},
  \institution{School of Biological Sciences, The University of Edinburgh}
  \city{Edinburgh}
  \country{UK}
}

\author{Michael U. Gutmann}
\orcid{0000-0002-5329-9910}
\email{Michael.Gutmann@ed.ac.uk }
\affiliation{%
  \institution{School of Informatics, The University of Edinburgh}, 
  \city{Edinburgh}
  \country{UK}
}


\begin{abstract}
The design of biological systems is hindered by uncertainty arising from both intrinsic stochasticity of biomolecular reactions and variability across laboratory or experimental conditions. In this work, we present a sequential framework to optimize genetic circuits under both forms of uncertainty. By employing simulator models based on differential equations or Markov jump processes alongside a reinforcement learning (RL) policy-based approach, our method suggests experiments that adapt to unknown laboratory conditions while accounting for inherent stochasticity.
While previous Bayesian methods address uncertainty through iterative experiment–inference–optimization cycles, they typically require computationally expensive inference and optimization steps after each experimental round, leading to delays. To overcome this bottleneck, we propose an amortized approach trained up-front across a distribution of possible uncertain parameters. This strategy sidesteps the need for explicit parameter inference during the design cycle, enabling immediate, observation-based adaptation.
We demonstrate our framework on models for heterologous gene expression and a repressilator circuit, showing that it efficiently handles both molecular noise and cross-laboratory variability. 

\end{abstract}

\ccsdesc[500]{Applied computing~Systems biology}
\ccsdesc[500]{Computing methodologies~Uncertainty quantification}
\ccsdesc[500]{Computing methodologies~Sequential decision making}
\ccsdesc[300]{Applied computing~Computational biology}
\ccsdesc[300]{Computing methodologies~Artificial intelligence}
\ccsdesc[300]{Computing methodologies~Control methods}
\ccsdesc[300]{Computing methodologies~Modeling and simulation}
\keywords{Automated, Sequential, Design, Genetic circuits, Amortization, Reinforcement, Policy-based, POMDP, Epistemic, Uncertainty}

\received{n/a}
\received[revised]{n/a}
\received[accepted]{n/a}

\maketitle

\section{Introduction}

The ability to design robust biological systems with specified functions is a cornerstone of engineering biology \cite{brophy2014principles}. Engineered systems have applications ranging from the production of therapeutic drugs \cite{trosset2015synthetic, keefer1981human} and industrial chemicals \cite{lee_comprehensive_2019} to environmental remediation \cite{jiang2025metabolic}. Reliable design of such functionalities is hindered by uncertainty about the biological system and the exact experimental conditions. A design that works well in one lab may fail in another, making it challenging 
to predict which design will achieve the desired outcome.

\begin{figure}[t] 
\centering 
\includegraphics[width=\linewidth]{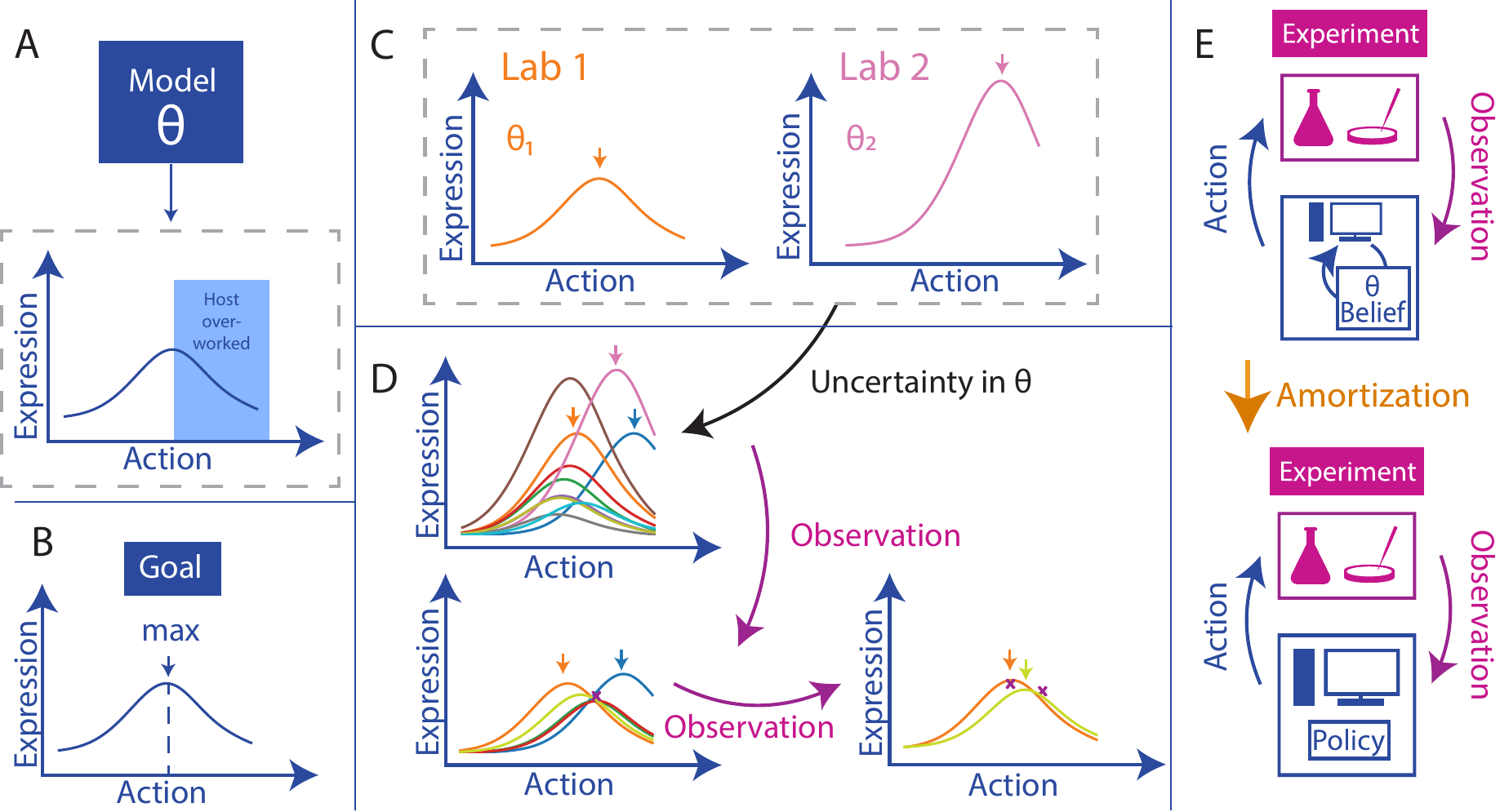} 
\caption{Overview of the approach using the example task of maximising gene expression in a biological system.
\textbf{A and B:} A parameterized model predicts the expected gene expression level in a biological system in response to an action and the goal is to find the action that maximizes gene expression.
\textbf{C:} Model parameters can be uncertain and vary between laboratories. Depending on the specific parameters, the optimal action may change. 
\textbf{D:} Different colored curves represent different possible values of $\theta$, i.e., the uncertainty, and arrows indicate the corresponding optimal actions. By performing an action and observing the result, uncertainty is reduced, ultimately improving knowledge of the optimal action. 
\textbf{E:} Traditional Bayesian approaches require updating the belief each time an observation is received, which can be computationally costly (illustrated by the inner loop on the top). In contrast, a policy-based approach can immediately recommend new actions after receiving an observation, effectively bypassing repeated inference and optimization steps. The policy’s goal is to sequentially adapt to unknown parameter values $\theta$ to ultimately achieve maximal expression. }
\Description{} 
\label{fig:overwiev}
\end{figure}

To guide the design of genetic systems, computational simulations based on mechanistic models are widely used \cite{brophy2014principles}. These models are typically formulated as systems of differential equations or as Markov jump processes (MJPs) \cite{gillespie1977exact} to capture intrinsic stochasticity. They allow experimenters to explore the behavior of genetic systems \emph{in silico} before performing costly wet-lab experiments.
To identify promising system designs, simulations are often combined with optimization strategies \cite{hiscock2019adapting,ma2009defining,li2017incoherent,woods2016statistical, merzbacher2023bayesian,qiao2019network,dasika2008optcircuit,otero2017automated,verma2021trade, filo2026genai}. 
As an example, consider the expression of a foreign gene, where one can control the promoter strength (action) and must balance increasing gene expression against overloading the host (see Figure~\ref{fig:overwiev}A). Optimization can then be applied to maximize the expected yield of the foreign protein (see Figure~\ref{fig:overwiev}B).
Such optimization approaches enable efficient exploration of the design spaces, but a key limitation lies in handling the uncertainty \cite{kobiela2026risk, sequeiros2023automated, schladt_automated_2021}.

Uncertainty can be broadly categorized into two types: aleatoric and epistemic. Aleatoric uncertainty reflects the intrinsic stochasticity of biological systems, e.g.\ molecular reactions, and is irreducible; it is present in every experiment and inevitably alters observed results. Epistemic uncertainty, by contrast, arises from incomplete knowledge of the system and the experimental conditions. 



Epistemic uncertainty is relevant in practice because, for instance, the true effective kinetic rates governing a biological system are rarely known precisely and can vary across experimental settings due to differences in growth conditions, reagent batches, or host strains. 
This uncertainty causes 
designs optimized under one set of 
parameters to fail or perform suboptimally when implemented under slightly different conditions (see Figure \ref{fig:overwiev}C). 
Epistemic uncertainty can be reduced through experimentation, which motivates  sequential design, where information from earlier experiments is used to refine predictions and guide the selection of subsequent designs. In some cases, a single iteration where a prior belief on parameter values is updated using experimental data, followed by optimization, is sufficient to obtain good designs \cite{kobiela2026risk, gerlach2020koopman}. But one single iteration is often not sufficient to disambiguate between several possibly optimal designs, and a sequence of experiments is needed to determine truly optimal one. This is illustrated in Fig.~\ref{fig:overwiev}D, where a single observation (the purple cross) does not sufficiently reduce the uncertainty to disambiguate the optimal design (arrows indicate possible locations). 

\begin{figure}[t] 
\centering 
\includegraphics[width=\linewidth]{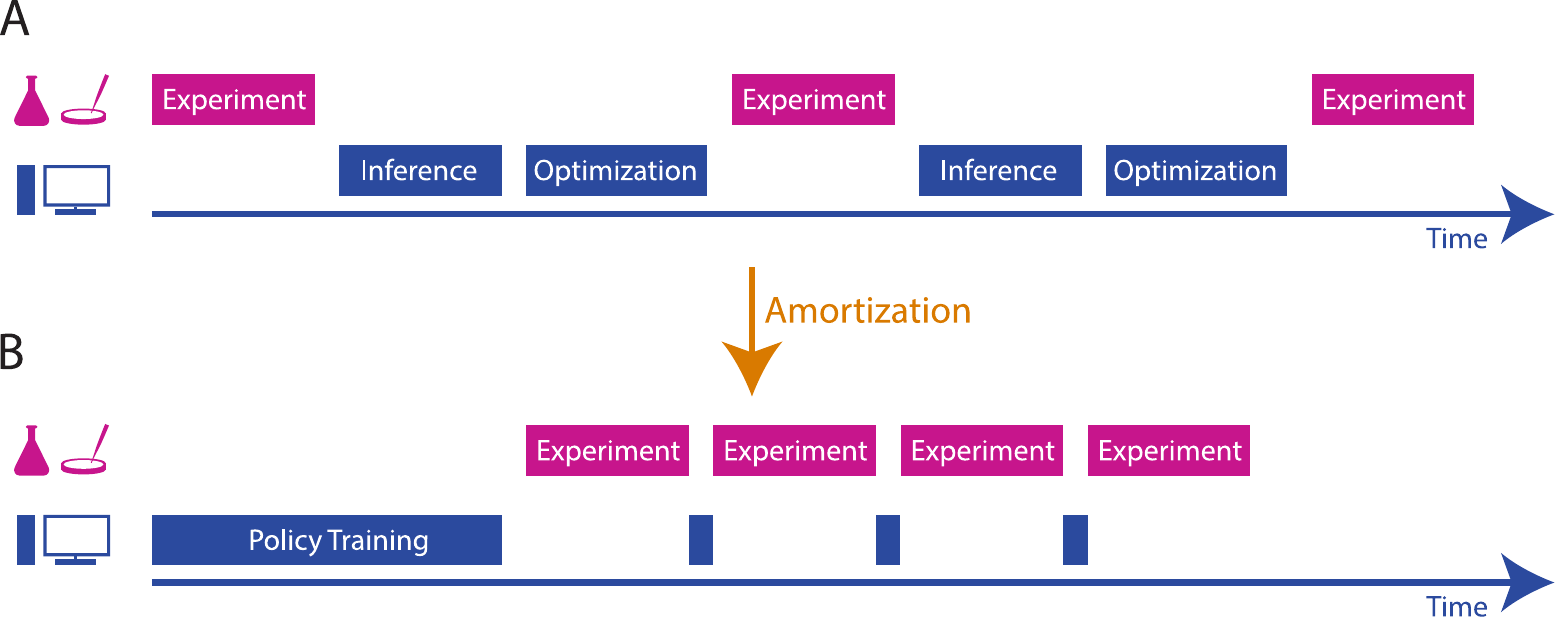} 
\caption{Timelines comparing inference-optimization based approach \cite{kobiela2026risk} (A) and amortized policy-based design (B). While the policy requires an initial time investment for training, it can be deployed immediately after each experiment produces an observation, unlike the traditional approach that must perform inference and optimization after every experiment.   .
}
\Description{} 
\label{fig:amort}
\end{figure}

A natural strategy is therefore to iterate this procedure: after each experiment, update the parameter belief using the newly collected data and re-optimize the design. While conceptually straightforward, this approach introduces computational overhead and delays between experimental cycles, as inference and optimization must be repeated after every observation (see Fig.~\ref{fig:overwiev}E). This not only results in delays between experiments (as illustrated in Fig.~\ref{fig:amort}), but may also require fine-tuning the inference and optimization algorithms during deployment to ensure that they work reliably for the specific problem at hand.

To address this limitation, we propose a naturally sequential alternative that handles uncertainty while eliminating the need for inference and optimization between experimental rounds. We cast experimental design as a partially observable Markov decision process (POMDP), where the true system parameters constitute a latent state that is specific to each laboratory setting, while actions are the controllable components of the system. Although training the policy may require substantial upfront computation, once deployed the algorithm can immediately select the next action after each observation is received (see Fig.~\ref{fig:amort}B). This is enabled by amortization: during training, the policy is exposed to many possible realizations of the model parameters sampled from the prior, allowing it to learn a direct mapping from experimental histories to actions.

We evaluate the proposed approach on two representative synthetic biology tasks: 
maximization of heterologous gene expression and tuning an oscillator toward a desired frequency. Using synthetic datasets, we examine how the learned policy adapts its experimental interventions over sequential rounds in the presence of latent environmental variation, without performing explicit parameter inference. We also investigate the role of prior knowledge by incorporating mechanistic models during training and compare the resulting behavior with uninformed optimization strategies such as Bayesian optimization.

\section{Previous work}
\begin{figure}[t]
  \centering
  \includegraphics[width=0.55\linewidth]{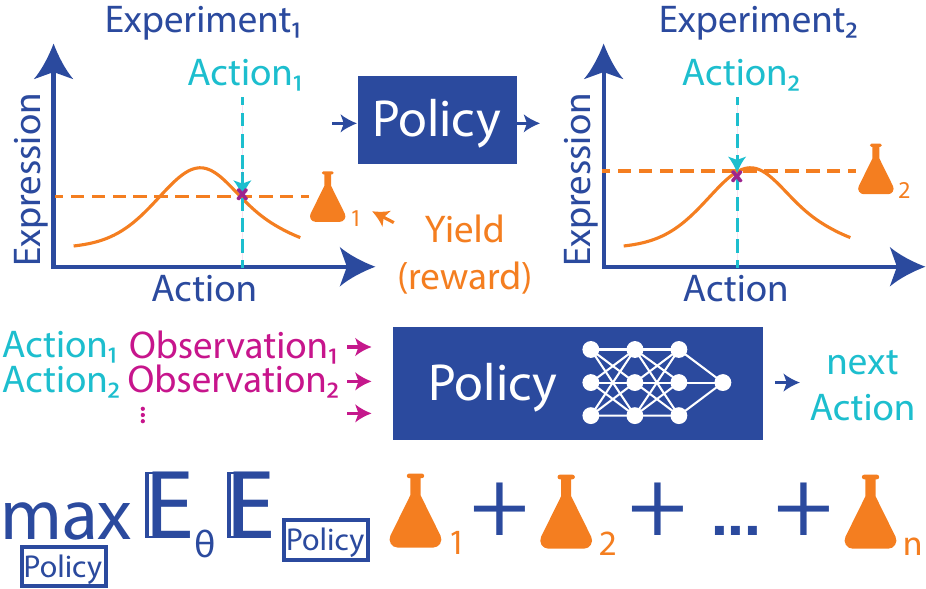}
  \Description{}
  \caption{ Overview of the methodology. We train a neural network policy that takes all previously tried actions and corresponding observations to recommend the next action. The training objective is to maximize the total yield across $n$ experiments by following the policy.
  }
  \label{fig:train_objective}
\end{figure}

Previous work applying reinforcement learning (RL) to synthetic biology has largely focused on bioreactor control~\cite{treloar2020deep} and optimal experimental design~\cite{treloar2022deep}, rather than genetic circuit design. In this latter area, early studies used RL with a discrete action space to design a deterministic dynamical system to obtain oscillatory dynamics from random initializations~\cite{giannantoni2023optimization}. RL played the role of an optimizer, with the policy operating on model-derived quantities (e.g.\ bifurcation properties). In our work, the model, in form of simulator, is used during training to learn a policy, but, at deployment, the policy operates directly on experimental observations without requiring access to the underlying model, enabling direct application in laboratory settings. This formulation is reflected in how actions and observations are defined. Rather than using model-derived quantities (e.g., bifurcation properties), we restrict observations to experimentally measurable outputs, such as gene expression trajectories. This allows the policy trained in simulation to transfer to the deployment phase where experimental data and not model-generated data is being used. We also consider design under both aleatoric and epistemic uncertainty. Aleatoric uncertainty arises from intrinsic stochasticity in biochemical reactions and measurement noise; in our framework, it is modeled either through observational noise or through Markov jump processes that explicitly capture stochastic reaction dynamics. We consider reducible epistemic uncertainty over model parameters, whose values are implicitly inferred from the full history of observations and actions that is directly provided as input to the policy (see Fig.~\ref{fig:train_objective}A), as opposed to the most recent observation.

Furthermore, treating the design parameters as actions allows the policy to select the initial design and explore the entire continuous design space, rather than operating on a discrete action space as e.g. in \citet{giannantoni2023optimization}. Consequently, the policy can leverage all available information to adapt its decisions in the presence of epistemic uncertainty from the very first experiments. This capability is particularly important during early stages of experimentation, where only a limited number of experiments can be performed and prior mechanistic knowledge encoded in the biological model can provide substantial guidance. While general-purpose optimization methods such as Bayesian optimization (BO; see, e.g., \citet{garnett2023bayesian}) may become attractive for larger experimental budgets, our empirical results show that, in early-stage experimentation, approaches that exploit prior biological knowledge achieve better-performing designs than methods that rely primarily on data collected during the optimization process.

Further contrasting with previous work, we note that conditioning decisions on histories of actions and observations is common in sequential Bayesian optimal experimental design (BOED) approaches~\cite{blau2022optimizing,fosterdeep,ivanova2021implicit,treloar2022deep} and in history-based approaches for handling partial observability in RL~\cite{NIPS2010_edfbe1af}. However, the objectives of BOED differ fundamentally from ours; those methods aim to maximize information gain about uncertain system parameters, and therefore may deliberately select experiments that are highly informative about the model but not necessarily useful for achieving a desired functional behavior. As a consequence, the resulting designs may not perform well with respect to the underlying engineering objective. In contrast, our approach directly optimizes for functional performance of the system. From the very beginning of the experimental process, the policy selects designs that aim to perform well under the latent system parameters while simultaneously gathering information that helps refine future decisions. This focus on performance-oriented experimentation aligns more closely with practical engineering goals, where the ultimate objective is to identify designs that reliably achieve a desired behavior rather than to fully characterize the underlying model.

\section{Methods}
\subsection{Genetic circuit design as a POMDP}
\label{gen_pompd}

We study the problem of optimizing the design of a genetic circuit when a simulator model is available that predicts system behavior as a function of a latent parameter vector $\theta$ (e.g., binding and dissociation kinetic rate constants) and controllable design parameters $a$ (e.g., promoter strengths or inducer concentrations).  

We formulate this problem as a Partially Observable Markov Decision Process (POMDP). Some background on POMDPs is provided in Appendix \ref{app:pomdp_details}, and Fig.~\ref{fig:graph} provides a graphical representation of our approach.

\begin{figure}[t]
  \centering
  \includegraphics[width=0.5\linewidth]{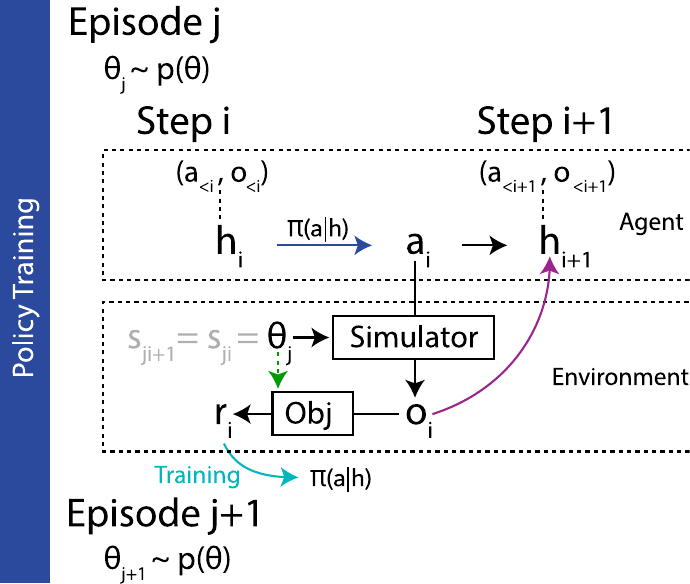}
  \Description{}
  \caption{The training loop. In each episode, different values of the simulator parameters $\theta$ are sampled from the prior distribution. At each step, the policy uses all previous actions and corresponding observations $h_i$ to select an action $a_i$. The simulator predicts the resulting observation $o_i$, and the reward function evaluates the outcome, generating a reward $r_{i}$. The reward function is computed solely from observations, except for the host-aware growth case study, where $\theta$ is used directly in the reward (green dotted arrow).}
  \label{fig:graph}
\end{figure}

The simulator defines a mapping
\begin{align}
o_t \sim \Omega(a_t, \theta),
\end{align}
where $o_t$ denotes the observable output of the system, such as a gene expression trajectory (e.g., via reporter gene fluorescence).  
The simulator may be stochastic and non-differentiable (e.g., when implemented via a Markov Jump Process).

We assume a prior distribution over latent parameters,
\begin{align}
\theta \sim p(\theta),
\end{align}
representing uncertainty about the biological system before experiments are performed.  
At the start of each episode, parameters are sampled as $\theta_j \sim p(\theta)$ and remain fixed throughout.

The latent state in the POMDP corresponds to these unknown parameters $\theta$.  
At each experimental iteration $t$, the agent selects design parameters $a_t$ and observes 
$
o_t \sim \Omega(a_t,\theta_j)$,
During deployment, simulated observations are replaced by experimental measurements.

We assume a task-specific reward function $R(a_t,\theta_j)$ that computes a scalar score $r_t$,
\begin{align}
r_t = R(a_t,\theta_j),
\end{align}
based on the action taken and the latent state of the environment. While this is the general case, the reward is often only computed from the observation $o_t$, e.g., the total expression yield as in Fig.\ \ref{fig:train_objective} or the deviation from a desired oscillation frequency in Fig.\ \ref{fig:repressilator}. 
During training, $\theta$ may also be used when defining the reward in certain case studies (see \ref{second_case}), but the policy itself does not observe $\theta$ or the reward. Instead, it conditions only on the history of actions and observations,
\begin{align}
h_t = (a_1, o_1, \dots, a_{t-1}, o_{t-1}),
\end{align}
and selects actions according to
\begin{align}
a_t \sim \pi_\psi(a_t \mid h_t),
\end{align}
where $\pi_\psi$ is a policy parameterised by $\psi$. This formulation ensures that the learned policy can be deployed in real laboratory settings where $\theta$ is unknown and only experimental observations are available.

We assume time, budget, or other practical constraints allow for a finite number of design iterations $T$. The goal is to sequentially select actions $a_1, \dots, a_T$ that achieve the design objective while accounting for both epistemic uncertainty in $\theta$ and intrinsic stochasticity in the simulator. For small $T$, biologically uninformed methods often fail to gather sufficient information, whereas our approach leverages the simulator to enable more sample-efficient design (see Section \ref{first_case}). Additional formal details are provided in Appendix~\ref{app:genetic_pomdp}.
\subsection{Policy optimization}

We optimize the policy using Proximal Policy Optimization (PPO)~\cite{schulman2017proximal}.  
The policy $\pi_\psi(a_t \mid h_t)$ maps the interaction history to a distribution over design parameters, while a value network $V_\phi(h_t)$ estimates the expected cumulative reward, when the actions of $\pi_\psi(a_t \mid h_t)$ are followed.

The objective is to maximize the expected return over an episode of $T$ experimental iterations,
\begin{align}
J(\psi) =
\mathbb{E}_{\theta_j \sim p(\theta),\, a_{1:T} \sim \pi_\psi}
\left[
\sum_{t=1}^{T} r_t
\right]. \label{training-objective}
\end{align} 

Because the simulator can be stochastic and non-differentiable, gradients cannot be computed by differentiating through it.  
Instead, the policy gradient estimator is used:
\begin{align}
\nabla_\psi J(\psi) =
\mathbb{E}\left[
\sum_{t=1}^{T}
\nabla_\psi \log \pi_\psi(a_t \mid h_t) \, \hat{A}_t
\right],
\end{align}
where $\hat{A}_t$ is an advantage estimate computed using the value network. The advantage is defined as 
\begin{align}
   A_\phi(h_t, a_t) := Q_\phi(h_t, a_t) - V_\phi(h_t), 
\end{align}
 where \(Q_\phi(h_t, a_t)\) is the expected cumulative return starting from history \(h_t\), taking action \(a_t\) at time \(t\), and thereafter following the policy, and 
 \(V_\phi(h_t)\) is the expected cumulative return from \(h_t\) under the policy. Intuitively, \(A_\phi(h_t, a_t)\) measures how much better taking action \(a_t\) is compared to the average performance of the policy from the same history. PPO further stabilizes learning through a clipped surrogate objective that prevents large policy updates.  
Training proceeds over many simulated episodes with parameters $\theta_j \sim p(\theta)$, allowing the policy to implicitly infer hidden system parameters from past observations and adapt experimental designs accordingly.
Additional algorithmic details are provided in Appendix~\ref{app:ppo_details}.

\section{Results}

\subsection{Maximizing heterologous gene expression with a host-aware simulator}
\label{first_case}

\begin{figure}[t]
  \centering
  \includegraphics[width=0.75\linewidth]{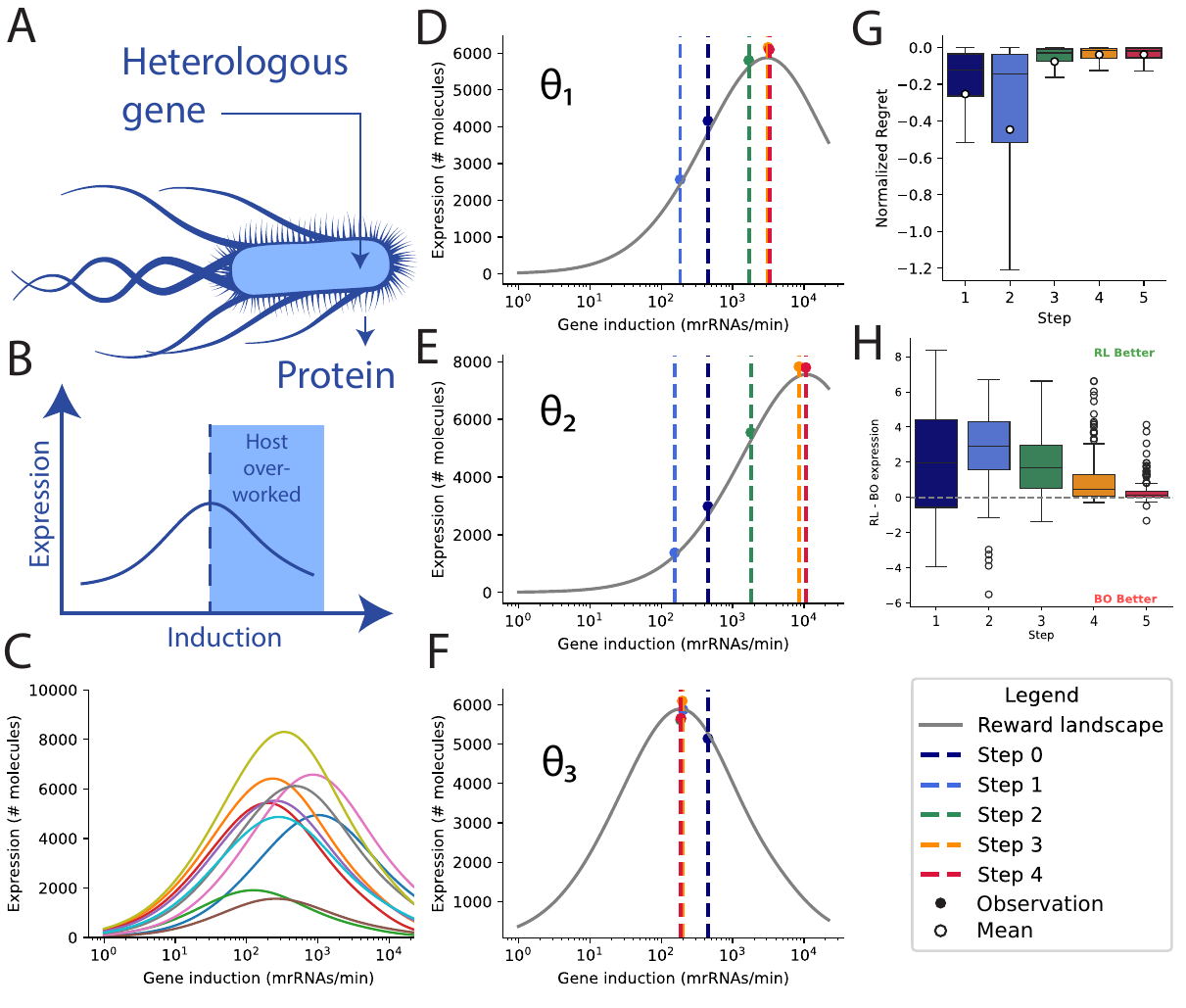}
  \caption{Design of heterologous gene expression system. A: The host organism \emph{E. coli} is inserted with foreign (heterologous) gene e.g. GFP. B: Example prediction of the model for fixed parameters $\theta$, it turns out that increasing gene induction can overload the organism and result in lower expression. C: Variability of induction (action) to expression (observation and reward) curves depending on the sample of $\theta$ from prior belief. D, E and F: The plot showing the actions taken by the policy for 5-step horizon, for three different test values  $\theta \sim p(\theta)$. Aggregated results for 1000 test values  $\theta \sim p(\theta)$, y-axis shows the normalized regret, i.e. difference between actual expression for each action and optimal expression. H: Comparison with a Bayesian optimization 
  baseline for 100 test samples $\theta \sim p(\theta)$.}
  \Description{}
  \label{fig:results_1}
\end{figure}

We consider the design of a gene expression system that produces a foreign protein in the \textit{Escherichia coli} bacterium, which is a central tasks in many synthetic biology applications \cite{nikolados2021prediction}.   
The objective is to tune the induction  of the expression system to maximize protein output while accounting for limited cellular resources and variability in host physiology, captured by the latent parameters $\theta$, see Figs.~\ref{fig:results_1}A and B.

The system is modeled using a host-aware mechanistic framework from \citet{weisse2015mechanistic} and \citet{nikolados2021prediction}, which explicitly models 
the dynamics of major proteome components $x \in \{r, t, m, q, \text{gfp}\}$. Here, $r$ denotes ribosomes, $t$ transcriptional machinery, $m$ metabolic proteins, $q$ housekeeping proteins, and $\text{gfp}$ the heterologous protein of interest. The model incorporates metabolism, transcription, ribosome binding, translation, dilution, degradation processes:

\begin{align}
\begin{array}{c c c}
\textbf{Metabolism} & \textbf{Transcription} & \textbf{Ribosome Binding} \\
s \xrightarrow{v_\text{imp}} s_\text{int} \xrightarrow{v_\text{cat}} \blue{n_s} e &
\varnothing \xrightarrow{w_x} m_x &
p_r + m_x\overset{\blue{k_{bx}}}{\underset{\blue{k_{ux}}}{\rightleftharpoons}} c_x \\[1em]
\textbf{Degradation} & \textbf{Translation} & \textbf{Dilution} \\ 
m_x \xrightarrow{d_{m,x}} \varnothing, \quad p_x \xrightarrow{d_x} \varnothing &
n_xe + c_x \xrightarrow{v_x} p_r + c_x + p_x &
e, m_x, c_x, p_x \xrightarrow{\blue{\lambda}} \varnothing
\label{eq:host-reactions}
\end{array}
\end{align}
for each protein $x \in \{r, t, m, q, \text{gfp}\}$.
Some of the rates are functions of other parameters and species:
\begin{alignat}{2}
v_{\text{imp}} &= \frac{p_t v_t s}{K_t + s}, 
&\quad v_{\text{cat}} &= \frac{p_m v_m s_i}{K_m + s_i},\\
w_x &= \frac{\blue{w_{x,\max}} \, e}{\theta_x + e}, \quad x \in \{r,t,m,\text{gfp}\}, 
&\quad w_q &= w_{q,\max} \frac{e}{\theta_q + e} \frac{1}{1 + (\rho_q / K_q)^{h_q}}, \\
v_x &= \frac{c_x}{n_x} \frac{\gamma_{\max} e}{e + K_\gamma}, 
&\quad \blue{\lambda} &= \frac{\gamma(e)}{M} \sum_{x \in \{r,t,m,q,\text{gfp}\}} c_x, \label{lambda} \\
M &= \sum_x n_x p_x + \sum_x n_r c_x, 
&\quad \gamma(e) &= \frac{\gamma_{\max} \text{e}}{e + K_\gamma.} 
\end{alignat}


In this model, the external nutrient $s$ is imported into the cell at rate $v_{\text{imp}}$, which depends on the abundance of transport proteins $p_t$ and follows Michaelis--Menten kinetics with maximum rate $v_t$ and half-saturation constant $K_t$. The internalised substrate $s_{\text{int}}$ is then catabolized into energy $e$ at rate $v_{\text{cat}}$ by metabolic enzymes $p_m$ with catalytic parameters $v_m$ and $K_m$, producing $n_s$ units of energy per substrate molecule. Gene expression is initiated by transcription, where mRNA $m_x$ is synthesized at rate $w_x$, which increases with energy availability $e$ and saturates according to $\theta_x$; for housekeeping genes $q$, transcription is further regulated by a negative feedback term involving the proteome fraction $\rho_q$, threshold $K_q$, and Hill coefficient $h_q$. Ribosomes $p_r$ reversibly bind mRNA to form translation complexes $c_x$ with association and dissociation rates $k_{bx}$ and $k_{ux}$, respectively. Protein synthesis occurs at rate $v_x$, which depends on the number of complexes $c_x$, the elongation rate $\gamma(e)$, and the protein length $n_x$, with elongation limited by energy through a saturating function with parameters $\gamma_{\max}$ and $K_\gamma$.

Both mRNA and proteins are subject to degradation with rates $d_{m,x}$ and $d_x$, respectively. In addition, all intracellular species are diluted by cellular growth at rate $\lambda$, which is computed from the translational activity across the proteome through Eq. (13). 
Together, these processes capture the interplay between metabolism, gene expression, and growth, and explicitly account for competition for shared resources such as ribosomes and energy. All default parameter values are given in \citet{nikolados2021prediction}, which are based on the original study by \citet{weisse2015mechanistic}.

The key parameters of interest in our case study are highlighted in \blue{blue} above. They are $w_{\text{gfp},\max}$, which controls induction of the heterologous protein; the nutrient efficiency $n_s$; and the ribosome–mRNA binding and unbinding rates, $k_b$ and $k_u$ of gfp, denoted without protein index for simplicity. In the next section we will also impose growth rate $\lambda$ constraints.

We built a simulator model based on Ordinary Differential Equations (ODE) derived from the reaction system above to predict heterologous gene expression 
as a function of the design variable (action) $w_{\text{gfp},\max}$ and the latent parameters $\theta = \{n_s, k_u, k_b\}$. The observation $o$ is the steady state expression of gfp, calculated by simulating for a sufficiently long time horizon using \emph{Rosenbrock23} method from DifferentialEquations.jl library \cite{rackauckas2017differentialequations}. This steady state value is also the reward, as we aim to maximize the heterologous protein yield. 
Priors over the latent parameters are specified as $n_s \sim \mathcal{U}(0,1)$, $k_b \sim \mathcal{U}(0,2)$, and $k_u \sim \mathcal{N}(1,1)$ truncated to the interval $(0,2)$. They are centered around the original 
values reported in \citet{nikolados2021prediction}. The Normal prior for $k_u$ is intended to emulate a scenario in which a parameter is, in principle, controllable and partially characterized but not precisely known due to implicit regulation and/or limited knowledge. 

For a fixed $\theta$, the model predicts that increasing induction beyond a critical threshold can overload the host, reducing heterologous expression due to resource competition (Fig.~\ref{fig:results_1}B). Sampling $\theta$ from the prior reveals substantial variability in the induction–expression relationship (Fig.~\ref{fig:results_1}C), demonstrating that both the location of the optimal action and the scale of the induction–expression curves can vary. To address this, we trained a policy capable of adapting the induction parameter $w_{\text{gfp},\max}$ to unknown values of $\theta$ to maximize the cumulative yield of the heterologous protein.

We trained the policy over a five-step experimental horizon. Fig.~\ref{fig:results_1}D–F shows the actions selected by the trained policy for three example latent parameter vectors drawn from the prior 
$\theta_i \sim p(\theta)$. The policy consistently selects the same initial action, reflecting the lack of prior data to inform the latent parameters at that point. Once the first observation becomes available, the policy begins to adapt to the true value of $\theta$, balancing exploration and exploitation in subsequent steps and generally converging toward the optimal action by the final step. Notably, while the policy is trained to maximize the cumulative protein yield rather than identify the optimal action, it turns out that the policy does discover the optimal action.

Aggregated performance over 1000 test samples of $\theta$ is quantified using the normalized regret, which quantifies the difference between the expression achieved by the policy and the optimal expression for the same system parameters, divided by the optimal expression value. Fig.~\ref{fig:results_1}G summarizes this metric across all test cases for the five-step experimental horizon. The low regret observed in the final three steps indicates that the recommended actions closely approach the optimal solution.

Finally, Fig.~\ref{fig:results_1}H compares our approach with a Bayesian optimization baseline (see Appendix \ref{appendix:bo}) evaluated on 100 test samples. We plot for each of those samples differences 
in protein expression obtained with both approaches, aggregated in a box plot. As the differences between our approach and BO tends to be positive for all steps. 
This comparison demonstrates that our method can efficiently leverage prior knowledge provided by the mechanistic model to recommend better actions compared to an uninformed approach. These results suggest that such an approach holds promise for guiding time-consuming or costly experiments, aiding the design process during early stages of experimentation.

\subsection{Maximizing heterologous gene expression with growth rate constraints}
\label{second_case}

\begin{figure}[h]
  \centering
  \includegraphics[width=1\linewidth]{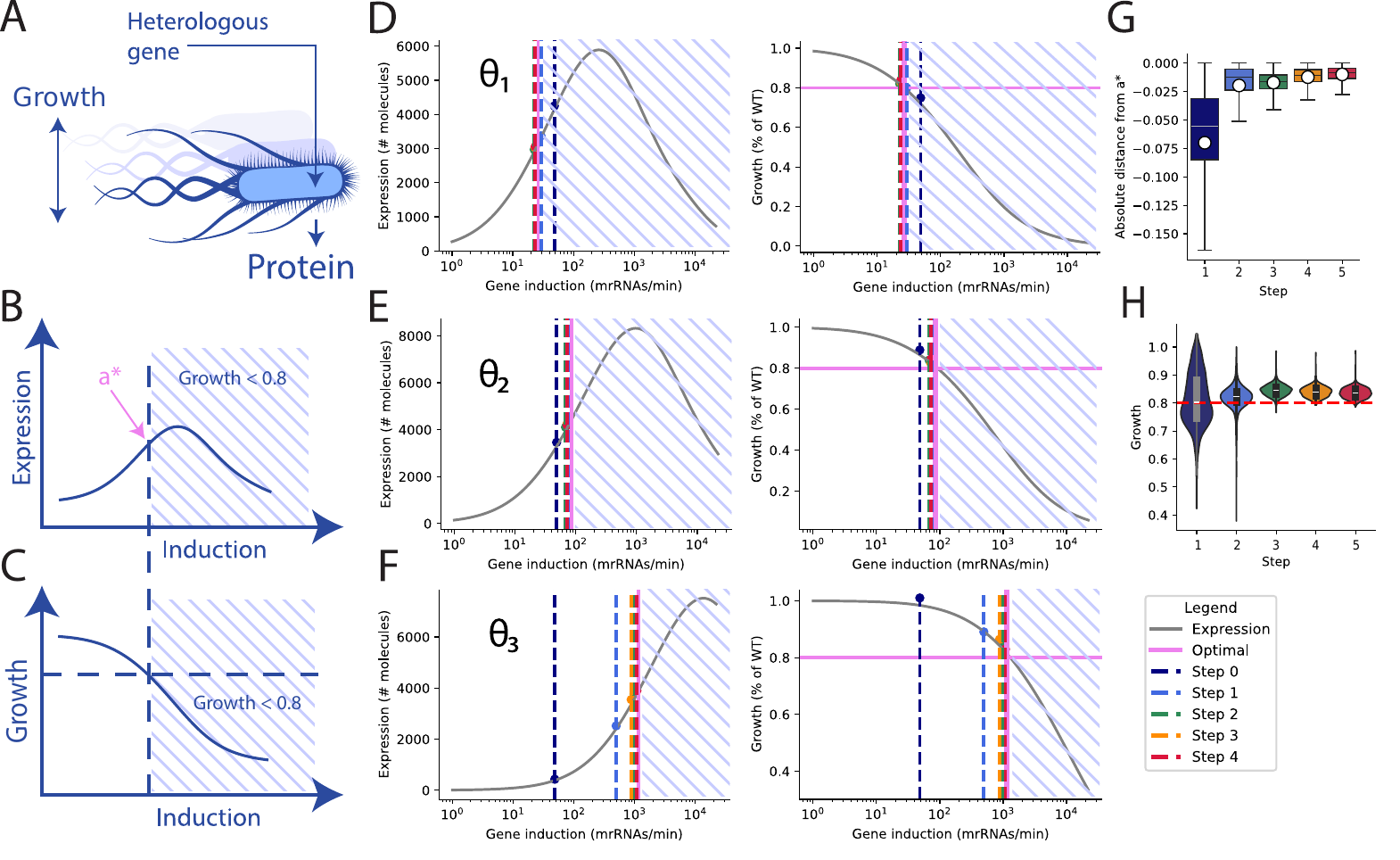}
  \caption{Host-aware heterologous gene expression with minimal impact.
\textbf{A:} The host-aware model can predict not only protein expression but also the growth of the host. Here, we assume that both expression and growth are observable.  
\textbf{B and C:} In this work, we assume that growth should not fall below a certain threshold (specifically 0.8). Therefore, the goal is to maximize expression outside the region where growth is too low, indicated by the shaded regions. The optimal action is denoted by $a^{*}$.  
\textbf{D, E, and F:} These plots show the actions taken by the policy over a 5-step horizon for three different test values $\theta \sim p(\theta)$, displaying both expression and growth. Depending on the uncertain parameters $\theta$, the location of the threshold can change, making the problem challenging. The optimal location can shift both due to differences in the position of the expression peak and variations in the threshold location.  
\textbf{G and H:} Aggregated results from 100 test samples are presented, showing both the distance from the optimal action and the growth value. Growth generally remains close to the threshold. 
}
  \Description{}
 \label{fig:results_2}
\end{figure}

We extend the previous study to account not only for heterologous protein production but also for host growth $\lambda$ (Fig.~\ref{fig:results_2}A), using the same mechanistic model (see Equation~\ref{lambda} for the definition of $\lambda$). In this scenario, both the steady state heterologous protein expression $o_\text{expr}$ and the host growth rate $o_\text{growth}$ are assumed observable and the action is the induction parameters $w_{\text{gfp},\max}$. The design objective is to maximize protein expression while ensuring that host growth does not fall below a predefined threshold, here set to 0.8. This constraint limits the feasible design space, excluding regions where growth is too low (shaded areas in Fig.~\ref{fig:results_2}B–C).

The optimal induction action $a^{*}$ is determined by balancing high protein expression against the growth constraint. Variability in latent host parameters $\theta$ shifts both the location of the expression peak and the effective growth threshold, making the optimization problem more problematic than unconstrained expression maximization (Fig.~\ref{fig:results_2}B–C).

To address this, we first trained a regressor that predicts the optimal action $a^{*}$ given a value of $\theta$. For the training of the regressor,  we sampled 100 values of $\theta$ from the prior and computed the corresponding induction-expression curves using the model. For each curve, $a^{*}$ was identified via grid search while taking into account the growth constraint, which is easily done since $a$ is a scalar. Next, we trained a policy using as reward function $R(a, \theta_j)$ the absolute distance between $a$ and the optimal action $a^{*}$ predicted from $\theta_j$ by the regressor. Note that this approach does not enforce that growth is strictly larger than the cutoff, just that it is close-by. Importantly, the history (which serves as the input to the policy) does not include rewards, and therefore $\theta$ is not available to the policy. Instead, the policy must 
sequentially adapt based only on observed outputs. The reward function is used solely during training via training objective (\ref{training-objective}) and is not employed during deployment (ensuring no leakege of \(\theta\)).

We trained the policy over a five-step experimental horizon. Fig.~\ref{fig:results_2}D–F illustrate the actions selected for three example latent parameter configurations $\theta \sim p(\theta)$, showing trajectories of both protein expression and host growth. The results demonstrate that the policy adapts induction levels to maintain growth above the threshold, even when the gene expression peak occurs in a region of low growth. Notably, the effective location of the growth constraint varies with $\theta$, highlighting the need for sequential adaptation to host-specific dynamics.

Aggregated results over 100 test samples of $\theta$ are shown in Fig.~\ref{fig:results_2}G–H. Fig.~\ref{fig:results_2}G quantifies the deviation from the optimal action, while Fig.~\ref{fig:results_2}H shows the corresponding host growth values. Across all cases, growth remains near or above the threshold. Overall, the method remains promising, effectively adapting induction levels to maximize heterologous protein expression even when constraints on host growth are imposed.


\subsection{Designing a genetic oscillator under stochastic and parametric uncertainty} \label{second_case}

\begin{figure}[t]
  \centering
  \includegraphics[width=0.89\linewidth]{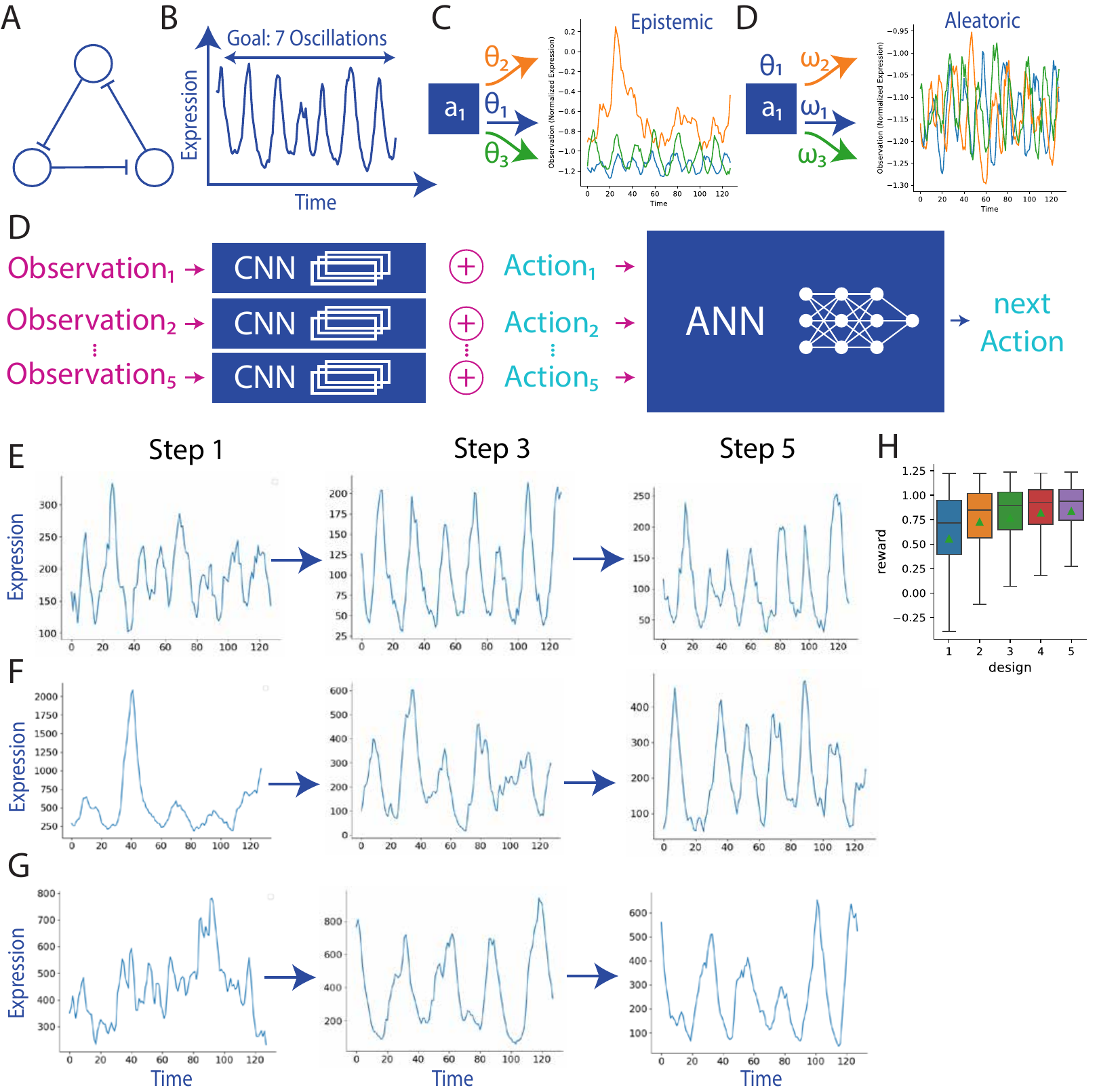}
  \caption{\textbf{A:} Schematic representation of the oscillator circuit -- repressilator. Three genes repress each other in a circular fashion.
\textbf{B:} The goal is to design an oscillator with a specific frequency.
\textbf{C:} The same action applied to different uncertain parameters can result in very diverse responses (epistemic uncertainty).
\textbf{D:} Even if the uncertain parameters are fixed to specific values, the same action can still result in highly variable responses depending on the irreducible randomness due to molecular noise, represented by different random seeds $\omega$. In this case, however, the frequency and amplitude are much less variable compared to epistemic uncertainty.
\textbf{E:} Neural network architecture of the policy. The observations are first processed with a shift-invariant convolutional layer to reduce phase dependency, which are then concatenated with corresponding actions and fed into a standard MLP, which outputs the next action.
\textbf{F,G,H:} Three examples of the design process, showing the observations at each step. In the first step, the policy does not receive any information yet and suggests a solution in the presence of both aleatoric and epistemic uncertainty. In later steps, observations are provided to the policy, reducing epistemic uncertainty and allowing the policy to recommend better actions. The final step does not always result in the desired frequency; however, it is much closer to the target, and oscillations are often more regular than in the initial designs.
\textbf{I:} Aggregated results showing how the reward function evolves for 100 test points of $\theta$. In each step, both the median (middle lines) and mean (triangles) improve.}
  \Description{}
  \label{fig:repressilator}
\end{figure}

We consider the design of a genetic oscillator based on the system in Fig.~\ref{fig:repressilator}A, in which three genes repress each other in a ring configuration. 
This circuit, termed the repressilator, has been shown to display oscillatory protein expression dynamics \cite{elowitz2000synthetic}. In this case study, the objective is to select design parameters that yield oscillations with a target 
frequency (Fig.~\ref{fig:repressilator}B), while accounting for both intrinsic stochasticity and uncertainty in system parameters.

As a preliminary step, we first verified that policy gradient methods can effectively optimize repressilator designs in the absence of parametric uncertainty. In this setting, the task reduces to optimizing circuit behavior under intrinsic stochasticity alone, similar to prior work \cite{sequeiros2023automated}. We found that our approach reproduces comparable solutions while remaining computationally efficient, converging in competitive time using only a single CPU thread (see Appendix \ref{app:no_epi} for details). This confirms that policy gradient methods provide a viable alternative to traditional optimization approaches in stochastic biochemical systems. Building on this result, we extended the setting to incorporate uncertainty over system parameters. To this end, the system is modeled using a stochastic reaction network adapted from \citet{sequeiros2023automated}, capturing transcription, translation, and degradation processes for each gene $i \in \{1,2,3\}$:
\begin{align}
\renewcommand{\arraystretch}{1.2}
\begin{array}{c c c c c}
& \textbf{mRNA Production} & \textbf{Protein Production} & \textbf{mRNA Degradation} & \textbf{Protein Degradation} \\ 
\textbf{Gene 1} & \emptyset \xrightarrow{ \alpha_1 } m_1 
& m_1 \xrightarrow{ k_X } m_1 + p_1 
& m_1 \xrightarrow{ \gamma_m } \emptyset 
& p_1 \xrightarrow{ \gamma_X } \emptyset \\
\textbf{Gene 2} & \emptyset \xrightarrow{ \alpha_2} m_2 
& m_2 \xrightarrow{ k_X } m_2 + p_2 
& m_2 \xrightarrow{ \gamma_m } \emptyset 
& p_2 \xrightarrow{ \gamma_X } \emptyset \\
\textbf{Gene 3} & \emptyset \xrightarrow{ \alpha_3 } m_3 
& m_3 \xrightarrow{ k_X } m_3 + p_3 
& m_3 \xrightarrow{ \gamma_m } \emptyset 
& p_3 \xrightarrow{ \gamma_X } \emptyset,
\end{array}
\end{align}
with $\alpha_1 = k_m \left( \epsilon + (1 - \epsilon) \frac{1}{1 + \left(\frac{p_3}{K}\right)^H} \right)$, $\alpha_2 = k_m \left( \epsilon + (1 - \epsilon) \frac{1}{1 + \left(\frac{p_1}{K}\right)^H} \right) $ and $\alpha_3 = k_m \left( \epsilon + (1 - \epsilon) \frac{1}{1 + \left(\frac{p_2}{K}\right)^H} \right)$.\\

The system is extended to a stochastic reaction network describing the coupled dynamics of transcription, translation, and degradation for three genes $i \in \{1,2,3\}$. Gene expression is modeled through repressing regulatory interactions, where mRNA production for each gene occurs at rate $\alpha_i$, with repression mediated by the protein product of another gene. Here, $k_m$ denotes the maximal transcription rate, $K$ is the repression threshold, $H$ is the Hill coefficient controlling the steepness of the regulatory response, and $\epsilon$ represents basal 
transcription that ensures a nonzero expression level even under full repression. Protein synthesis follows translation reactions of the form $m_i \xrightarrow{k_X} m_i + p_i$, where each mRNA acts catalytically to produce protein $p_i$ at rate $k_X$. mRNA and proteins are linearly diluted by cell growth 
with rate constants $\gamma_m$ and $\gamma_X$, respectively. The full system is simulated as a Markov jump process, thereby capturing intrinsic (aleatoric) stochasticity arising from the probabilistic nature of molecular reaction events \cite{elowitz2002stochastic}. We used \emph{Catalyst.jl} \cite{CatalystPLOSCompBio2023} and \emph{JumpProcesses.jl} \cite{rackauckas2017differentialequations} to derive and simulate the system form the above reactions using stochastic simulation algorithm (SSA) \cite{gillespie1977exact}.

In addition to this intrinsic noise, the model incorporates parametric (epistemic) uncertainty by treating the parameter vector $\theta = \{H, \gamma_X, \gamma_m, \epsilon\}$ as unknown but bounded, with prior distributions reflecting limited prior knowledge: $H \sim \mathcal{U}(3,7)$, $\gamma_X \sim \mathcal{U}(0.8,1.1)$, $\gamma_m \sim \mathcal{U}(4,50)$, and $\epsilon \sim \mathcal{U}(0.05,0.15)$. The design variables (actions) are given by $a = \{k_X, k_m, K\}$, which are selected by the policy within predefined ranges:
\begin{align*}
k_X \in [100, 1000], \quad
k_m \in [3, 120], \quad
K \in [10, 200].
\end{align*}


The simulator generates stochastic gene expression trajectories conditioned on the chosen design \(a\) and latent parameters \(\theta\). The observation \(o\) consists of the time series of protein concentration (we use \(p_1(t)\)). From this trajectory, we estimate the oscillation frequency  \(f(o)\) by identifying the location $\tau$ of the second peak of the normalized autocorrelation function \(C(\tau)\). The reward function is defined as the weighted sum of two terms:
\[
R(a, \theta) = -\big(f(o) - f^\star\big)^2 + \lambda \, C(\tau_2),
\quad \text{with } \lambda = 0.3.
\]
The first terms penalises the deviation of \(f(o)\) from the target frequency \(f^\star\). The second term is the value of the second peak, as used by \citep{sequeiros2023automated}; it
encourages more regular and sustained oscillatory behavior of the design.

To improve training stability, we sampled 500 different values of the uncertain latent parameters \(\theta\) and use them to normalize (rescale) the loss. This ensures that the optimization objective is well-balanced across parameter variability and prevents certain regions of the latent space from dominating the gradients.

Sampling $\theta$ from the prior reveals substantial variability in oscillatory behavior (Fig.~\ref{fig:repressilator}C), demonstrating that both the achievable frequency and the regularity of oscillations depend strongly on the underlying parameters. Even for a fixed $\theta$, repeated simulations exhibit variability due to intrinsic noise (Fig.~\ref{fig:repressilator}D).

To achieve the desired design goal in presence this kind of uncertainty, we trained a sequential decision-making policy that adapts to unknown values of $\theta$ while accounting for stochastic observations. Observations were processed using a convolutional neural network with shift-invariant layers to ensure phase invariance, and the extracted features are concatenated with the corresponding actions before being passed to a fully connected network that outputs the next action (Fig.~\ref{fig:repressilator}D).

We trained the policy over a 5-step experimental horizon. Fig.~\ref{fig:repressilator}E–G shows the actions selected by the policy for three example latent parameter configurations $\theta \sim p(\theta)$. As in the previous case study, the policy initially selects the same action due to the absence of information about $\theta$. Once the first observation becomes available, the policy begins to adapt, progressively reducing epistemic uncertainty and refining its recommendations. While early actions may produce irregular or off-target oscillations, later steps yield trajectories that more closely match the desired frequency and exhibit more regular behavior. In this case, due to broad priors, although the final designs improve upon the initial ones, it may not be possible to completely match the frequency. In the appendix, we present a simplified case study in which it is more feasible to achieve the target frequency across the prior range. We also provide additional results demonstrating robustness to random initializations, along with comparisons to oracle policies informed by the ground-truth parameters, which allow us to assess how well the policy adapts relative to the optimal value.

Aggregated performance over 100 test samples of $\theta$ is summarized in Fig.~\ref{fig:repressilator}H. We report the reward value at each step, showing consistent improvement in both mean and median performance. These results demonstrate that sequential, closed-loop design enables robust control of stochastic genetic circuits, effectively handling both intrinsic noise and uncertainty in system parameters, and extending the applicability of model-based design to realistic biological systems.

\section{Discussion}
In this paper, we present an reinforcement learning approach to sequentially optimize genetic circuits under epistemic and aleatoric uncertainty.
The case studies presented here illustrate that amortized sequential design policies can provide an effective and practical approach for optimizing biological systems under uncertainty, as demonstrated using synthetic data. Compared to more traditional approaches that iterate experiment-inference-optimization blocks, our approach avoids repeated inference and optimization during the experimental phase, decoupling the computational effort from the experimental process. Once trained, the policy can propose design actions in real time, circumventing potentially costly and computationally difficult steps of Bayesian inference and optimization.


One limitation to our study is the amortization gap: the policy may not perfectly match the optimal design for every possible latent parameter configuration. One way to mitigate this is through semi-amortized approaches, similar in spirit to work \cite{ivanova2024step} in Bayesian optimal design, where the policy is updated sequentially using posterior information. While this can improve adaptation to the true system, it comes at the cost of additional computation, as inference must be performed and the policy potentially retrained. Exploring such semi-amortized strategies is an interesting avenue for future work.

Another important consideration is that all experiments here were conducted using synthetic data. While synthetic evaluations allow controlled validation of the approach, real experimental systems may introduce additional sources of model mismatch or unmodeled dynamics. Handling potential model mismatch robustly is therefore an important future direction, particularly in biological settings where simulator fidelity may be limited. 

Our approach provides a natural framework for the low-data, uncertain scenarios, as they can combine prior knowledge and simulation-based inference with sparse observations. Furthermore, policy gradient methods offer an alternative to standard optimization approaches for designing genetic circuits under molecular noise such as \citet{sequeiros2023automated} when there is no epistemic uncertainty. This can be particularly useful in systems like the repressilator, where Markov Jump Process simulations are relatively fast but intrinsic noise is high, making traditional optimization challenging. Because policy gradient methods are local and do not require simulator gradients, they likely can scale to high-dimensional design spaces.

Overall, our results suggest that amortized sequential design provides an attractive tool, enabling rapid decision-making while maintaining flexibility to handle stochasticity, parametric uncertainty, and limited experimental data, while utilizing prior knowledge in form of the mechanistic model. Future work addressing model mismatch, semi-amortized updates, and validation in real experimental systems will further strengthen the applicability of this approach.

\section*{Code Availability}
The code, data, and trained models used in this study are publicly available at:
 \url{https://github.com/MichalKobiela/GeneCircuitsRL}
.
\begin{acks}
This work was supported by the United Kingdom Research and Innovation (grant EP/S02431X/1), UKRI Centre for Doctoral Training in Biomedical AI at the University of Edinburgh, School of Informatics. For the purpose of open access, the author has applied a creative commons attribution (CC BY) licence to any author accepted manuscript version arising.

During the preparation of this work the authors used ChatGPT in order to increase the readability of the text. After using this tool/service, the authors reviewed and edited the content as needed and take full responsibility for the content of the publication.

\end{acks}

\bibliographystyle{ACM-Reference-Format}
\bibliography{sample-base}

\appendix

\section{Methods details}
\subsection{Formal POMDP background}
\label{app:pomdp_details}

A Partially Observable Markov Decision Process (POMDP)~\cite{NIPS2010_edfbe1af}
is defined by the tuple
$(\mathcal{S}, \mathcal{A}, \mathcal{O}, T, \Omega, R, \rho_0, \gamma)$, where

\begin{itemize}
    \item $\mathcal{S}$ is the set of latent states $s \in \mathcal{S}$,
    \item $\mathcal{A}$ is the set of actions $a \in \mathcal{A}$,
    \item $\mathcal{O}$ is the set of observations $o \in \mathcal{O}$,
    \item $\rho_0(s)$ is the initial state distribution,
    \item $T(s_{t+1} \mid s_t, a_t)$ is the state transition model,
    \item $\Omega(o_{t+1} \mid s_{t+1}, a_t)$ is the observation model,
    \item $R(s_t, a_t, s_{t+1})$ is the reward function,
    \item $\gamma \in [0,1]$ is the discount factor.
\end{itemize}

At the start of an episode the latent state is sampled as
\begin{align}
s \sim \rho_0(\cdot).
\end{align}

At each timestep $t$, the agent selects an action $a_t$, after which the environment evolves according to
\begin{align}
s_{t+1} &\sim T(\cdot \mid s_t, a_t), \\
o_{t+1} &\sim \Omega(\cdot \mid s_{t+1}, a_t), \\
r_{t+1} &= R(s_t, a_t, s_{t+1}).
\end{align}

Because the latent state is not directly observable, decisions are conditioned on the action–observation history
\begin{align}
h_t = (a_0, o_1, a_1, o_2, \dots, a_{t-1}, o_t),
\end{align}
and actions are sampled from a policy
\begin{align}
a_t \sim \pi(a_t \mid h_t).
\end{align}
\subsection{Genetic circuit design as a POMDP}
\label{app:genetic_pomdp}

In our setting the latent state $a$ corresponds to the vector of model parameters $\theta$.
At the beginning of each episode $j$, parameters are sampled from a prior
\begin{align}
\theta_j \sim p(\theta).
\end{align}
These parameters remain fixed during the entire episode, reflecting that the underlying biological system does not change between experimental iterations. Consequently,
\begin{align}
s_t = \theta_j \qquad \forall t
\end{align}
and the transition model becomes deterministic,
\begin{align}
T(s_{t+1} \mid s_t, a_t) = \delta(s_{t+1}-s_t).
\end{align}

In the standard POMDP formulation the observation distribution is written as
\begin{align}
o_{t+1} \sim \Omega(\cdot \mid s_{t+1}, a_t).
\end{align}

However, since $s_{t+1}=s_t=\theta_j$ in our problem, the observation depends only on the current action and the fixed latent parameters.  
We therefore adopt the equivalent but simpler indexing convention
\begin{align}
o_t \sim \Omega(\cdot \mid \theta_j, a_t),
\end{align}
where $o_t$ denotes the experimental outcome produced by action $a_t$. Under this convention (and by starting indexing from 1 rather than 0), the interaction history becomes
\begin{align}
h_t = (a_1, o_1, \dots, a_{t-1}, o_{t-1}).
\end{align}
Similarly, the reward is defined with a shifted index as
\begin{align}
r_t = R(\theta_j, a_t),
\end{align}
which evaluates how well the observed circuit behavior satisfies the design objective. Shifting the reward index is commonly done in some frameworks, e.g.\ \citet{SpinningUp2018}.

\subsection{Policy optimization with PPO}
\label{app:ppo_details}

We optimize the policy using Proximal Policy Optimization (PPO)~\cite{schulman2017proximal}.  
The policy $\pi_\psi(a_t \mid h_t)$ is parameterized by $\psi$ and maps the interaction history to a distribution over actions, while a value network $V_\phi(h_t)$ estimates the expected cumulative reward from history $h_t$.

The expected return of a policy is
\begin{align}
J(\psi) =
\mathbb{E}_{\theta_j \sim p(\theta),\, a_{1:T} \sim \pi_\psi}
\left[
\sum_{t=1}^{T} r_t(\theta_j, a_t)
\right].
\end{align}
Because the simulator may be stochastic and non-differentiable, gradients of the reward with respect to actions cannot be computed directly.  
Instead, PPO uses the policy gradient estimator
\begin{align}
\nabla_\psi J(\psi) =
\mathbb{E}\left[
\sum_{t=1}^{T}
\nabla_\psi \log \pi_\psi(a_t \mid h_t) \hat{A}_t
\right],
\end{align}
where $\hat{A}_t$ denotes an estimate of the advantage function, typically computed using generalized advantage estimation \cite{schulman2015high}.

To stabilize training, PPO maximizes the clipped surrogate objective
\begin{align}
L^{\text{CLIP}}(\psi) =
\mathbb{E}
\left[
\min
\left(
r_t(\psi)\hat{A}_t,
\text{clip}(r_t(\psi),1-\epsilon,1+\epsilon)\hat{A}_t
\right)
\right],
\end{align}
where
\begin{align}
r_t(\psi) =
\frac{\pi_\psi(a_t \mid h_t)}
{\pi_{\psi_{\text{old}}}(a_t \mid h_t)}
\end{align}
is the probability ratio between the updated and previous policies.
Training proceeds by generating trajectories from the simulator using parameters sampled from $p(\theta)$ and updating the policy and value networks using stochastic gradient ascent on the surrogate objective $L^{\text{CLIP}}(\psi)$.

\subsection{Neural network architecture for the genetic oscillator}

The history is encoded as a stacked sequence of past actions and observations:
\begin{align}
x_t = [o_1, a_1, o_2, a_2, \dots, o_N, a_N].
\end{align}
Each $obs_i$ is a protein expression trajectory simulated with the MJP, and each $a_i$ is a vector of design parameters.  
This structure enables the policy to capture dependencies between past designs and outcomes and infer latent $\theta_j$.

The policy and value networks share a CNN-based feature extractor.  
Each observation $o_i$ is processed with a one-dimensional CNN and flattened:
\begin{align}
f_i = \text{CNN}(o_i),
\end{align}
then concatenated with the corresponding action $a_i$ to form
\begin{align}
z_t = [f_1, \dots, f_N, a_1, \dots, a_N].
\end{align}
This embedding is fed to fully connected layers to produce the final feature for PPO.

\subsection{Implementation and training}

Training uses \texttt{Stable Baselines3} \cite{stable-baselines3} PPO with Gaussian policies.  
The CNN feature extractor has latent dimension 128.  
Training runs for millions of steps, corresponding to thousands of simulated design--experiment iterations, allowing the policy to progressively infer hidden $\theta_j$ from history $h_t$ and adapt its design strategy (Fig.~\ref{fig:train_objective}D). Stable Baselines is designed for policies that are conditioned on a single observation. In our setting, however, the policy needs access to a sequence of past observations and actions. To handle this, at each timestep we construct an “effective observation” by stacking the history into a fixed-size representation and padding it with zeros where necessary. This zero-padded history is then passed to the policy.

For visualization, we employed Seaborn \cite{Waskom2021} and Matplotlib \cite{Hunter:2007} within the Python programming language. For the simulations, we utilized the Julia programming language \cite{Julia-2017}, together with the packages DifferentialEquations.jl (with JumpProcesses.jl) \cite{rackauckas2017differentialequations}, and Catalyst.jl \cite{CatalystPLOSCompBio2023}. For regressor training, we employed PyTorch \cite{paszke2019pytorch}.

\section{Genetic oscillator with biomolecular noise only---no epistemic uncertainty}
\label{app:no_epi}

In this experiment, we replicate the repressilator case study from \citet{sequeiros2023automated}, but instead of relying on a mixed-integer optimization framework and partial integro-differential equations, we employ reinforcement learning (RL) for design optimization and Markov jump processes (MJPs) for simulation. The objective is to identify parameter values that maximize the location of the second peak of the normalized autocorrelation function, used as a proxy for oscillation robustness.

\begin{figure}[h!]
    \centering
    \includegraphics[width=0.4\linewidth]{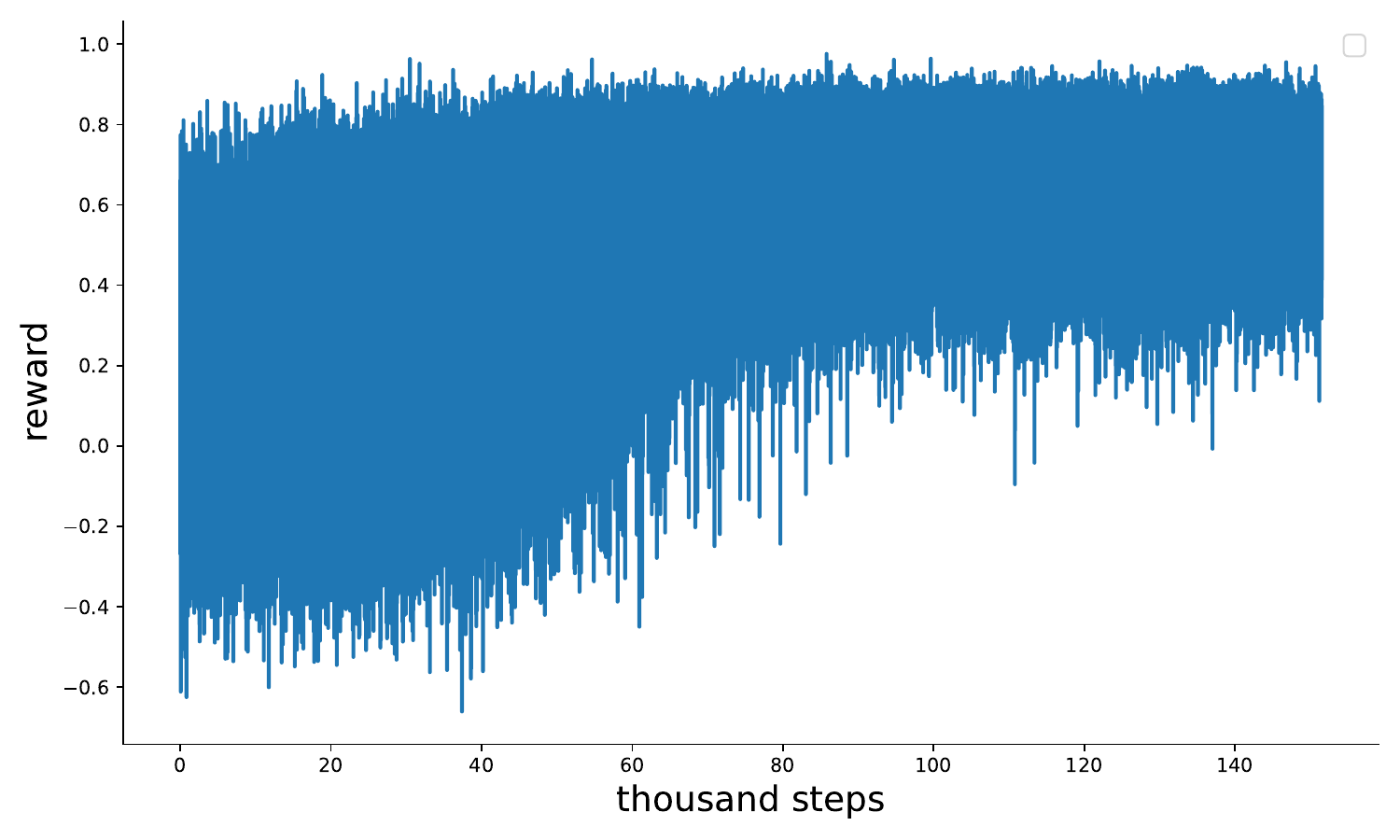}
    \includegraphics[width=0.4\linewidth]{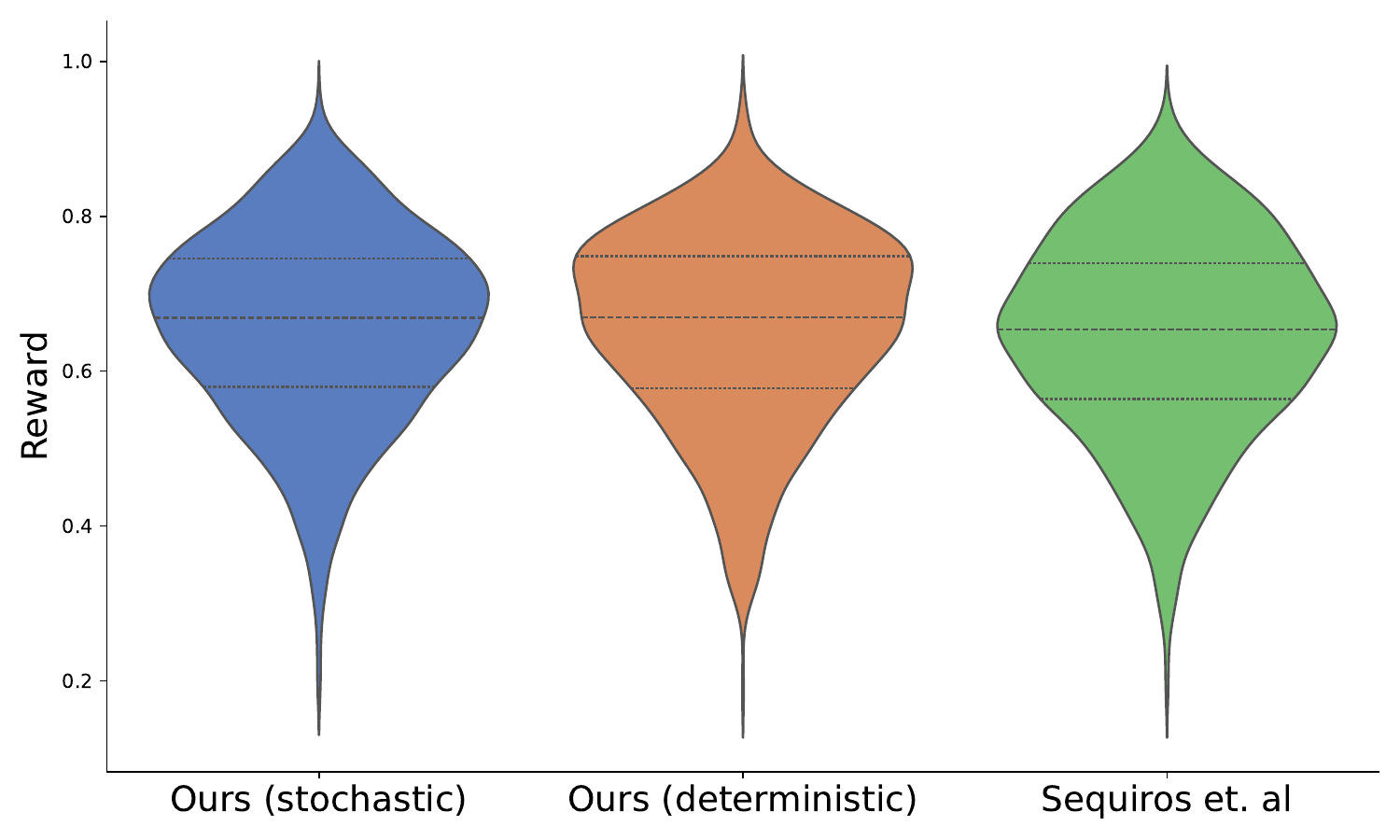}
    \includegraphics[width=0.8\linewidth]{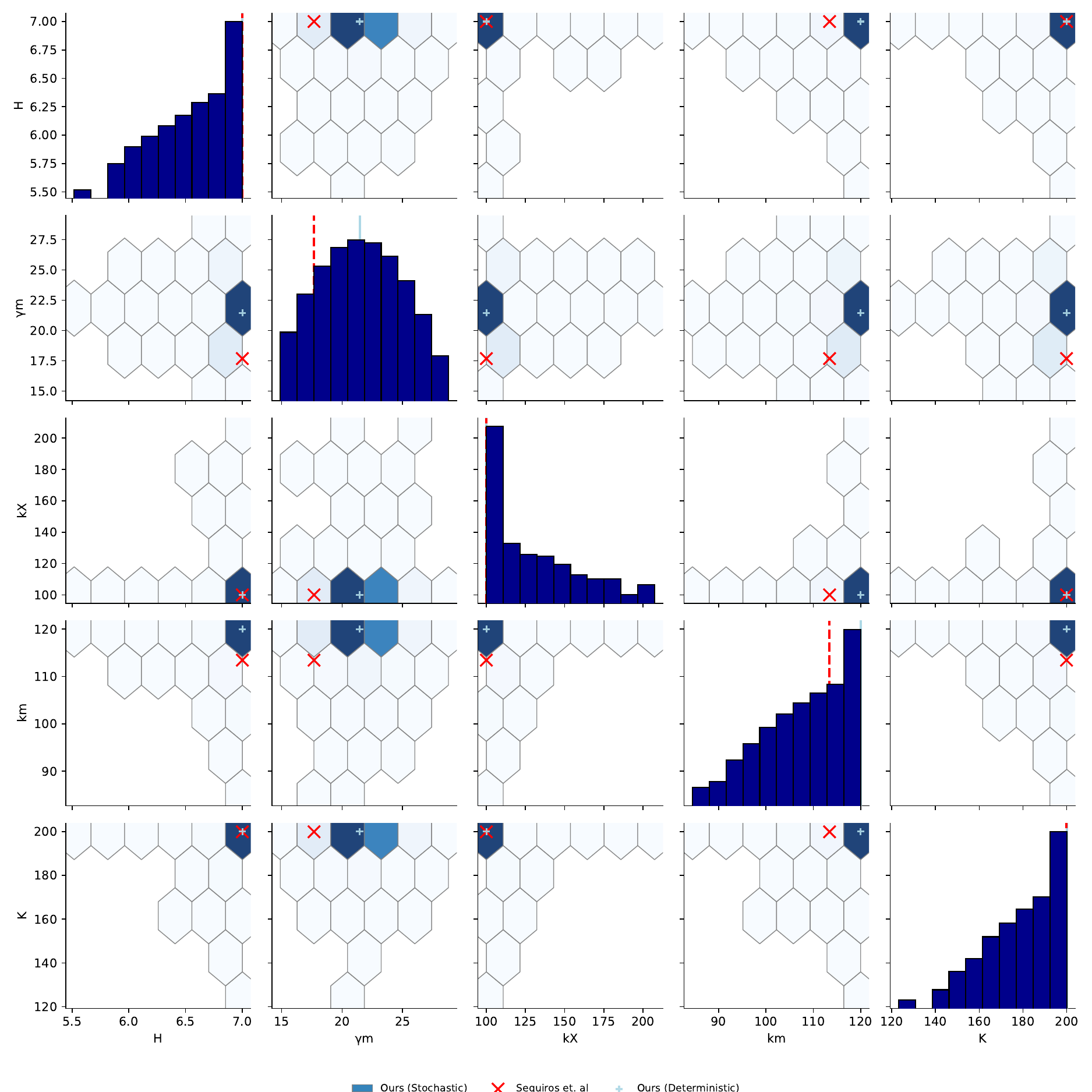}
    \caption{RL-based optimization of the repressilator under biomolecular noise. Top left: Training curve showing improvement of the reward function over time. Top right: Comparison between our stochastic policy, its deterministic mean, and the design reported by Sequeiros et al., showing comparable distributions of the reward. Bottom: Heatmap of sampled designs from the trained policy, its mean and Sequiros et. al design, which is contained within the policy.}
    \label{fig:first}
\end{figure}

Using RL, we obtained a policy that includes the optimal design reported by \citet{sequeiros2023automated}, as shown in Figure~\ref{fig:first}. Moreover, the distribution of the reward under our learned policy closely matches the distribution obtained using their optimal design. Since our policy is Gaussian, it is also possible to use its mean as a deterministic design instead of sampling. We found that this mean-based design achieves performance comparable to the one reported by \citet{sequeiros2023automated}. Furthermore, our approach required only 46 minutes on a single CPU thread, while \cite{sequeiros2023automated} reported 5.5 hours using a GPU. This demonstrates that approaches based on the score derivative trick, such as Proximal Policy Optimization (PPO), can effectively optimize genetic oscillators under biomolecular noise simulated via MJPs. While such simulations are non-differentiable and thus incompatible with standard gradient-based optimization methods, reinforcement learning provides a viable alternative by enabling efficient optimization in this setting.

\section{Bayesian optimization baseline} \label{appendix:bo}

We compare our approach against a Bayesian Optimization (BO) baseline implemented with a Gaussian Process (GP) surrogate model. To make the comparison fair we use the mechanistic model to inform the hyperparameters of the Gaussian process. Namely, we generated 100 input-output curves by sampling latent parameters $\theta \sim p(\theta)$ and evaluating the simulator at 20 equally spaced action-values for each sample $\theta$. The empirical mean of all observations was computed and subtracted from the data prior to fitting, resulting in a zero-mean dataset; this value was retained as the GP prior mean. The GP was equipped with a radial basis function kernel. The signal variance, length scale, and observation noise variance were estimated by maximizing the marginal likelihood on the generated dataset.

BO was run over a five-step horizon. The first action was selected uniformly at random. The next three actions were obtained by maximizing an acquisition function using the default \texttt{gp\_hedge} strategy in \texttt{scikit-learn} \citep{scikit-learn}, which randomly alternates between Probability of Improvement, Expected Improvement, and Lower Confidence Bound. The final action was selected as the maximizer of the GP posterior mean. Evaluation was performed over 100 independently sampled latent parameter settings $\theta \sim p(\theta)$ using different random seeds. For each instance, we recorded the objective values obtained by both methods and computed their differences.

\section{Additional results for the genetic oscillator}
\begin{figure}[h]
    \centering
    \includegraphics[width=0.65\linewidth]{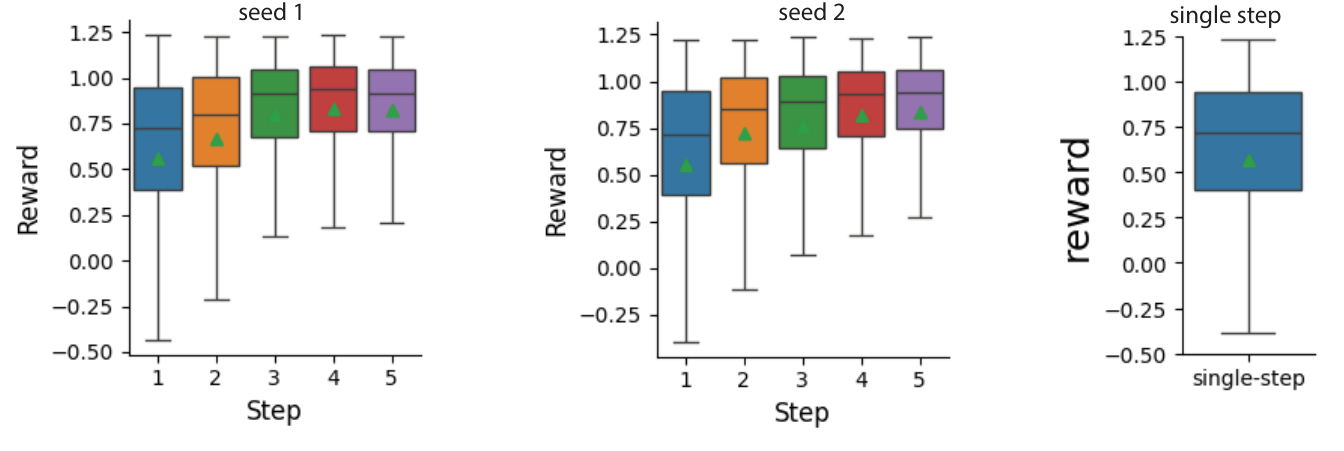}
    \caption{We present the differences between two random seeds used to initialize the policy and environment. Although the results are not identical, their performance is very similar. We also compare these to a single-step policy that maximizes the reward marginalized over the prior. This comparison shows that the first step of the multi-step policy achieves performance comparable to the solution obtained when the prior is marginalized out. This suggests that the policy is not merely finding progressively better solutions in a marginal sense (i.e., averaged over the prior) across steps, but is instead adapting to the unknown values of the uncertain parameters, see also Figure \ref{fig:omni}.}
    \label{fig:seed}
\end{figure}

Figure~\ref{fig:seed} illustrates the effect of different random seeds used to initialize the policy and environment. It also compares the multi-step policy to a single-step policy that optimizes the reward marginalized over the prior, highlighting their close agreement in the first step and supporting the interpretation that the policy adapts to the observations rather than only improving performance in a marginal (prior-averaged) sense across steps.

\begin{figure}[h]
    \centering
    \includegraphics[width=0.75\linewidth]{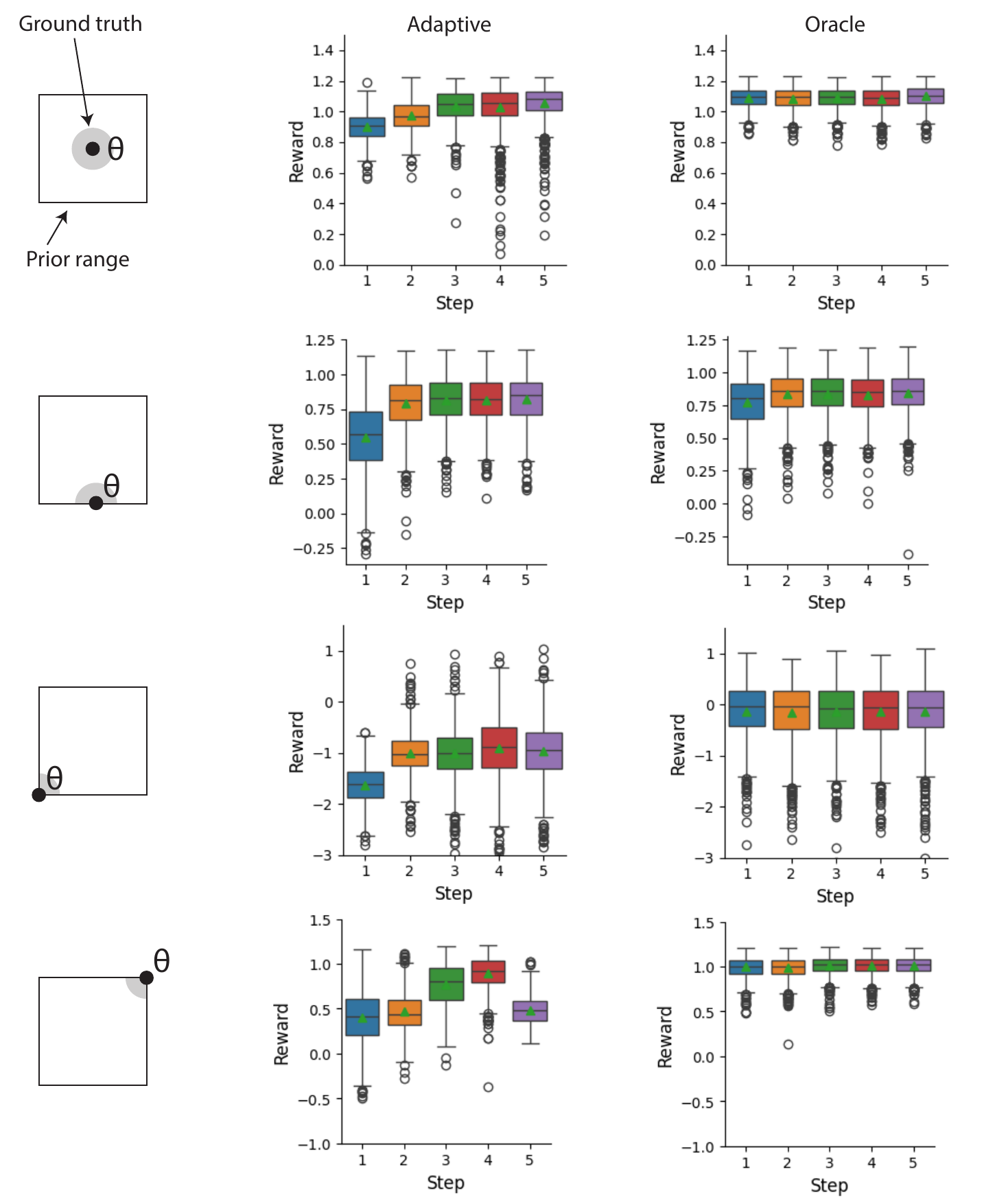}
    \caption{We compare the performance of the adaptive policy to four ``oracle'' policies, i.e., policies trained using the ground-truth parameter values rather than a prior distribution. We evaluate these across four different ground-truth settings of $\theta$: one at the center of the normalized parameter space, $(0,0,0,0)$; one on the boundary but not at a corner, $(1,-1,0,0)$; and two at the corners, $(-1,-1,-1,-1)$ and $(1,1,1,1)$. Note that $\theta$ is four dimensional; the figures show a two-dimensional representation.
Overall, the adaptive policy discovers solutions that are close in performance to the corresponding oracle policies, with the exception of the corner case $(-1,-1,-1,-1)$. This discrepancy is likely due to the amortization gap, as the volume of the surrounding neighborhood intersected with the parameter space is smallest at the corners (grey shaded areas); in the discussion we suggest possible remedies.
Importantly, the oracle policies do not always achieve fully functional designs (i.e., rewards close to 1), indicating that for some values of $\theta$, perfectly optimal designs may not be attainable.}
    \label{fig:omni}
\end{figure}

Figure~\ref{fig:omni} shows a comparison between the adaptive policy and four ``oracle'' baselines trained with access to ground-truth parameter values $\theta$. Across representative settings of $\theta$, the adaptive policy performs comparably to these oracle baselines, with some degradation in extreme regions of the parameter space. The figure further highlights that even the oracle policies do not always achieve fully functional designs, reflecting the inherent difficulty of the task.

\section{Genetic oscillator: Simplified case study}

In this case study, we consider a simplified version of the repressilator case study with more control over the parameters and less epistemic uncertainty. Specifically, here we take the parameter $\gamma_X$ to be controllable within the interval $[100,1000]$, the leakage is fixed to $\epsilon = 0.05$ and protein degradation rate is fixed to $\gamma_x=1.0$. The setup is otherwise the same as in the main case study.

\begin{figure}[h!]
    \centering
    \includegraphics[width=0.45\linewidth]{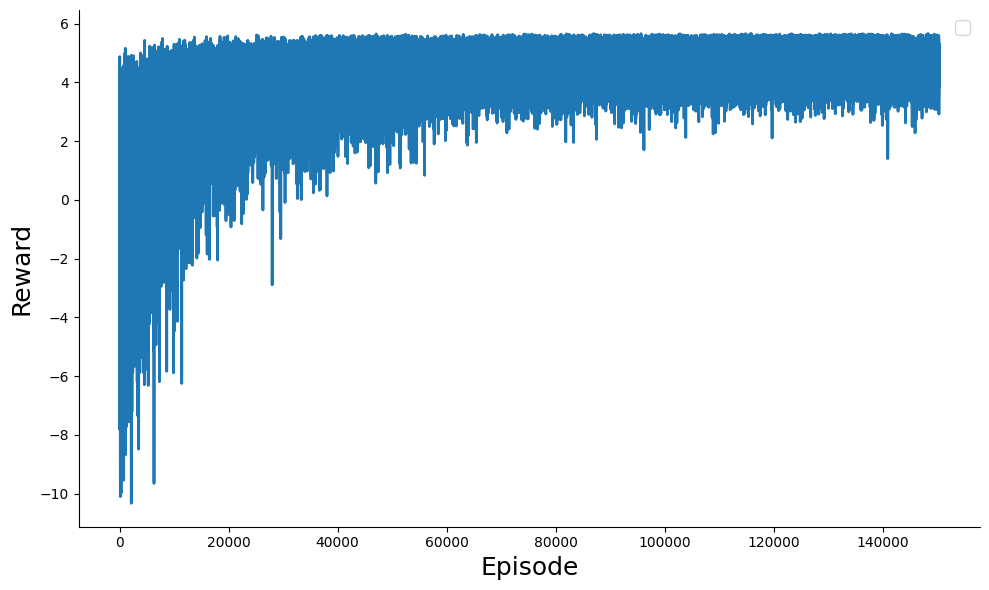}
    \includegraphics[width=0.45\linewidth]{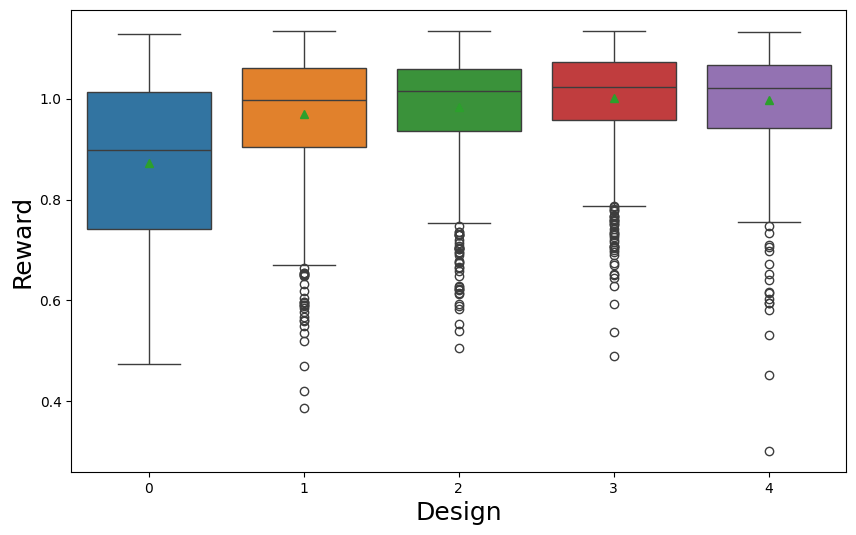}
    \includegraphics[width=0.9\linewidth]{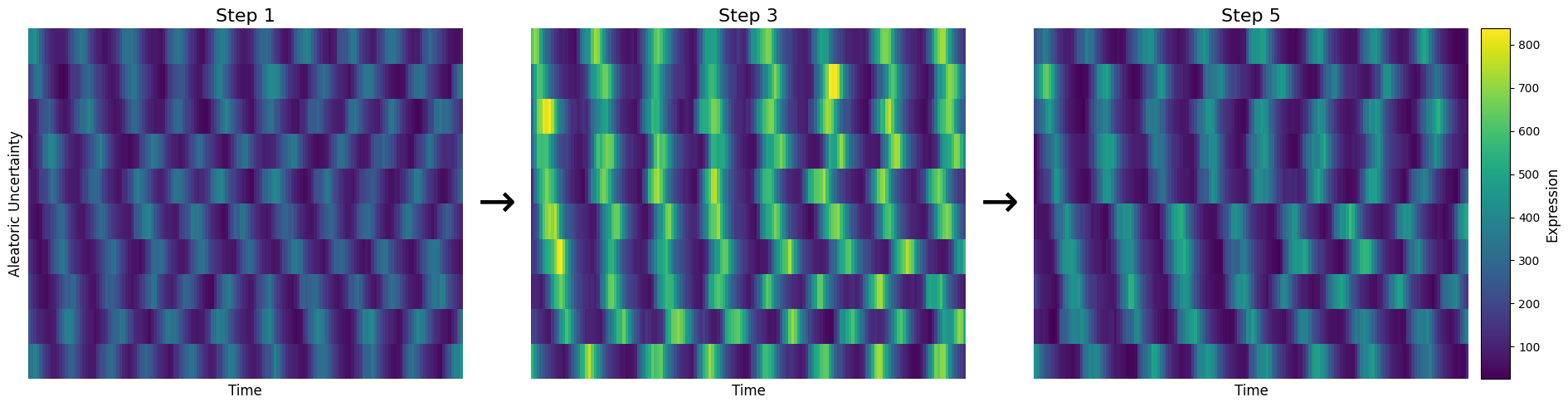}
    \includegraphics[width=0.9\linewidth]{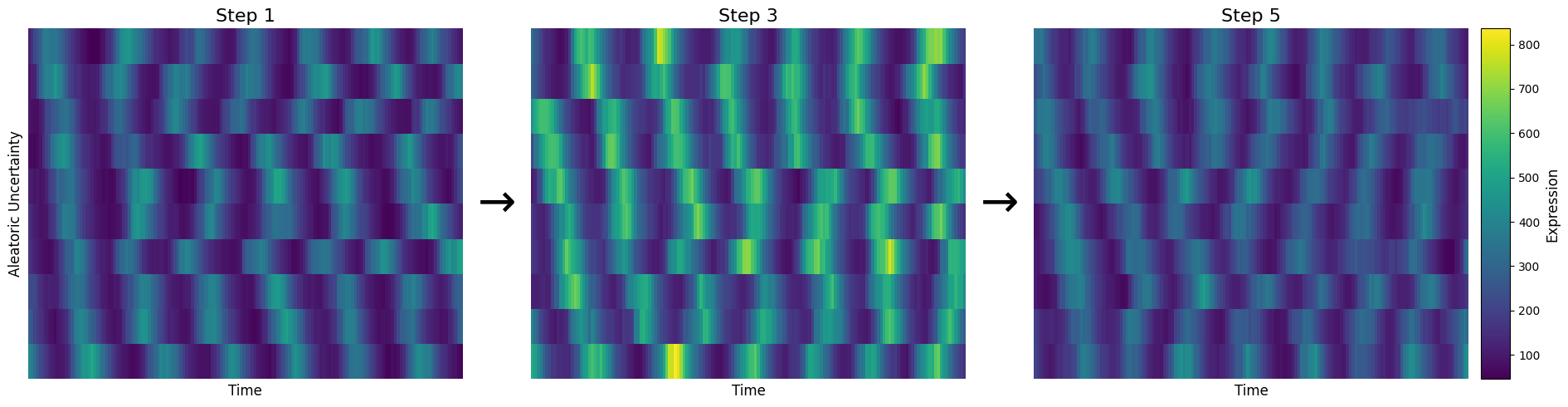}
    \caption{Simplified repressilator study. The policy adapts to unknown components of the system to achieve seven oscillations. Top left: Final reward increases as the policy is trained. Top right: Reward for each design in the sequential setting (design 1: before any observations; design 2: after observing first outcome, etc.). Middle: Heatmaps showing gene expression for the first, third, and fifth designs under one realization of the uncertain parameters. Bottom: Same as middle, but under a different realization of $\theta$.}
    \label{fig:thrid}
\end{figure}

Figure~\ref{fig:thrid} summarizes the results. The top-left panel shows the learning curve, indicating that the final reward improves steadily as the policy is trained over multiple episodes. The top-right panel plots the reward for each design step in the sequential setting. Unlike in the main case study, the final reward is concentrated around 1, where the desired number of oscillations is generally reasonably achieved. The middle and bottom rows present heatmaps of gene expression levels under specific values of the uncertain parameters (which are unknown to the policy). These figures demonstrate how the policy progressively adapts its strategy: for instance, in the middle row, the first design produces too many oscillations, but the policy adjusts its suggestions in subsequent steps and converges to a design yielding the desired target of seven oscillations. 
\end{document}